\documentclass{article}

 \usepackage[main, final]{neurips_2025}

\usepackage[utf8]{inputenc} % allow utf-8 input
\usepackage[T1]{fontenc}    % use 8-bit T1 fonts
\usepackage{hyperref}       % hyperlinks
\usepackage{url}            % simple URL typesetting
\usepackage{booktabs}       % professional-quality tables
\usepackage{amsfonts}       % blackboard math symbols
\usepackage{nicefrac}       % compact symbols for 1/2, etc.
\usepackage{microtype}      % microtypography
\usepackage{xcolor}         % colors

\usepackage{amsmath}
\usepackage{amssymb}
\usepackage{booktabs}
\usepackage{algorithm}
\usepackage{algorithmic}
\usepackage{graphicx}
\title{OOPMAS: Object-Oriented Multi-Agent Systems for Query-Level Workflow Generation}

\author{
  Qi Cheng \\
  Rutgers University\\
  New Brunswick, NJ\\
  \texttt{q.cheng@rutgers.edu} \\
  \And
  Shengyu Chen \\
  University of Pittsburgh \\
  Pittsburgh, PA \\
  \texttt{shc160@pitt.edu} \\
  \And
  Wei Cheng \\
  NEC Labs America\\
  Princeton, NJ \\
  \texttt{weicheng@nec-labs.com} \\
  \And
  Yiqun Xie \\
  University of Maryland\\
  College Park, MD \\
  \texttt{xie@umd.edu} \\
  \And
  Xiaowei Jia \\
  Rutgers University\\
  New Brunswick, NJ\\
  \texttt{xj159@cs.rutgers.edu} \\
  \And
  Haoyu Wang \\
  NEC Labs America\\
  Princeton, NJ \\
  \texttt{haoyu@nec-labs.com} \\
  \And
  Haifeng Chen \\
  NEC Labs America\\
  Princeton, NJ \\
  \texttt{haifeng@nec-labs.com} \\
}

\begin{document}

\maketitle

\begin{abstract}
Multi-agent systems (MAS) powered by large language models have shown strong performance across code generation, mathematical reasoning, and question answering. However, existing methods for automating MAS design mostly operate at the task level, producing a single fixed workflow per benchmark that is applied uniformly to all queries. This assumption fails under realistic conditions. Query difficulty varies widely within a task, and real-world workloads mix heterogeneous task types. We introduce OOPMAS, a training-free framework that generates both the agent set and the coordination workflow at the granularity of individual queries. Agents are represented as object-oriented class definitions with dedicated roles, tools, and persistent state, and workflows are expressed as executable main functions over these agent objects. A dynamic skill library accumulates structured lessons from execution feedback across optimization rounds, enabling in-context improvement without any gradient updates or fine-tuning. On a mixed-task benchmark of queries spanning code, math, and QA, OOPMAS achieves 89.6\% accuracy, outperforming the strongest baseline by 18.1 percentage points. A model-swap study across four LLM backbones shows consistent scaling, reaching 92.4\% with the strongest model.
\end{abstract}

\section{Introduction}

\begin{figure}[t]
    \centering
    \includegraphics[width=0.8\linewidth]{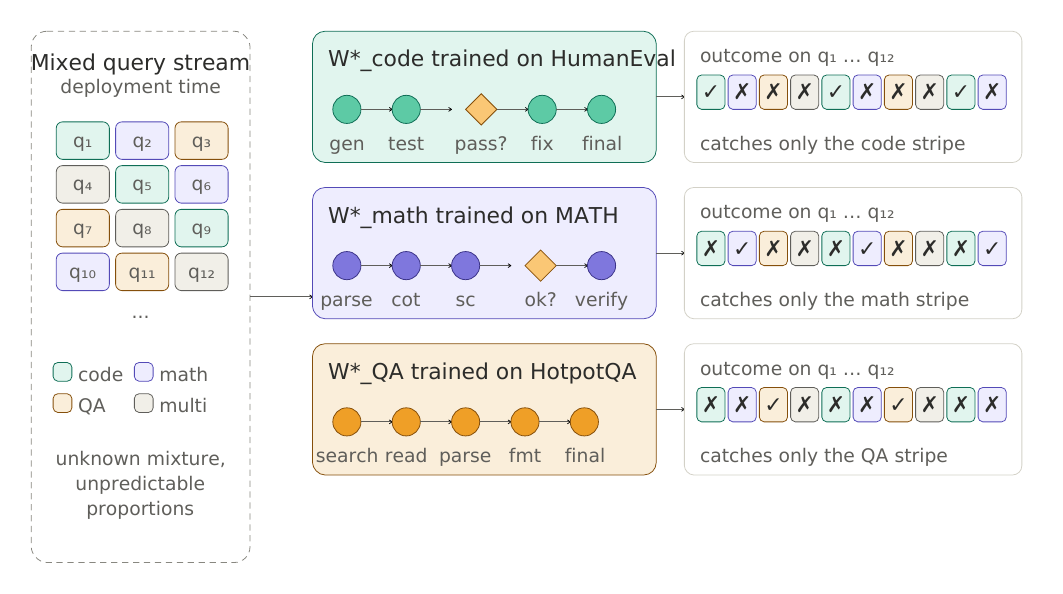}
    \caption{Task-level workflows fail on real-world query streams. Each fixed workflow catches only its native domain; queries from other domains and multi-domain queries (gray) are missed. No single workflow handles the full mixed distribution.}
    \label{fig:task-level-limitation}
\end{figure}

Multi-agent systems (MAS) powered by large language models have emerged as a leading paradigm for solving complex tasks across code generation, mathematical reasoning, question answering, and scientific analysis~\citep{metagpt,autogen,factualitysurvey,camel}
% Large language models (LLMs) have become the backbone of increasingly capable AI systems across code generation, mathematical reasoning, question answering, and scientific analysis~\citep{humaneval,gsm8k,hotpotqa,metagpt}. To reliably solve complex tasks, researchers have moved beyond single model queries toward multi-agent systems (MAS), in which multiple LLM-powered agents with distinct roles, and tools collaborate through structured workflows~\citep{factualitysurvey,autogen,metagpt,camel}.
Designing these systems, however, remains a largely manual process \citep{medprompt,cot,selfrefine,multipersona}. Each new task requires developers to decide which agents to create, what roles and tools to assign, and how agents should coordinate. This manual effort limits both scalability and adaptability. 

Recent works are moving to automate the generation and optimization of multi-agent workflows \citep{maas, adas, aflow, metaagent}. These methods have shown strong results on individual benchmarks including HumanEval~\citep{humaneval}, GSM8k~\citep{gsm8k}, MATH~\citep{math}, and HotpotQA~\citep{hotpotqa} by optimizing a single agentic workflow for each of the benchmarks. This one workflow per task approach share a common assumption that a single fixed workflow is the optimal solution uniformly to all queries within that benchmark. We argue that this task-level assumption breaks down in realistic deployment settings for two reasons (illustrated in Figure~\ref{fig:task-level-limitation}). First, within a single task category, the optimal MAS structure varies with query difficulty. An easy coding problem can be solved by a straightforward workflow, while a repository-level coding task benefits from a substantially more complex workflow. A single task-level workflow must commit to one topology and cannot adapt. Second, real-world query streams are heterogeneous mixtures of task types with unpredictable distribution. A workflow optimized for code generation provides no benefit when a math reasoning or question answering query arrives while supporting mixed workloads under the task-level paradigm would require maintaining a separate optimized system per domain plus a routing mechanism. Our argument is supported by the evaluation of these task-level methods on a mixed-task benchmark containing code, math, and QA queries, where the single optimized workflow often performs well on one task but poorly on others.

A second structural limitation for both task-level and query-level methods. Existing approaches search for workflow while holding the agent pool fixed \citep{aflow, adas, flowreasoner}. We argue that constraining the agent set limits the reachable solution space. If no agent in the pool possesses the domain expertise required by a particular query, no workflow in the search space over that pool can produce an optimal solution. The agents themselves must also be generated dynamically.

In this paper, we introduce OOPMAS (Object-Oriented Programming Multi-Agent Systems), a training-free framework that generates both the agent set and the workflow for individual queries. Given a query, a frozen LLM acting as a meta-agent first generates a pool of specialist agents with tailored role prompts, tool bindings, and model configurations, then produces an executable workflow program that orchestrates these agents to solve the query. Agents are represented as object-oriented class definitions, a choice that naturally encapsulates each agent's role, tools, and persistent state, and that aligns with code patterns prevalent in LLM pretraining data, leading to more reliable generation. The workflow is iteratively refined across multiple optimization rounds using a dynamic skill library that accumulates structured lessons from past successes and failures, enabling in-context improvement without any gradient updates, distillation, or fine-tuning.

We evaluate OOPMAS on two benchmarks. The standard \emph{Mixed} benchmark draws $120$~queries from six standard sources (HumanEval, MBPP, MATH, GSM8K, HotpotQA, DROP); the \emph{Hard Mixed} benchmark draws $180$~competition-level queries (AIME 2025/2026, LiveCodeBench, CodeContests, GPQA Diamond) with diverse output formats. OOPMAS achieves $89.6\%$ on Mixed (vs.\ $71.5\%$ for ADAS, $+18.1$~points) and $90.8\%$ on Hard Mixed (vs.\ $72.3\%$ for ADAS, $+18.5$~points). The same gap is preserved despite the much harder problem distribution, showing that per-query generation generalises rather than relying on benchmark-specific memorisation. A model-swap study across four LLM backbones (GPT-4.1 through GPT-5.2) shows that OOPMAS scales consistently with model capability, reaching $92.4\%$ with GPT-5.2 without any system-level changes.

Our contributions are as follows:
\begin{enumerate}
    \item \textbf{Problem formulation.} We formalise MAS generation as a query-level optimisation problem over a joint agent-and-workflow search space, and argue that both query-level adaptation and dynamic agent generation are necessary for handling heterogeneous task distributions.
    \item \textbf{OOPMAS framework.} We propose a training-free, two-phase pipeline consisting of dynamic OOP-style agent generation and iterative workflow optimisation with a structured skill library. The system requires no reinforcement learning, distillation, or fine-tuning and is immediately deployable on arbitrary task distributions.
    \item \textbf{Mixed and Hard Mixed benchmarks.} We introduce two benchmarks that test MAS generation under realistic heterogeneous workloads, including a competition-difficulty Hard Mixed benchmark spanning AIME, LiveCodeBench, CodeContests, and GPQA. OOPMAS substantially outperforms existing task-level and query-level baselines across both benchmarks and all domains, with the gap preserved as task difficulty grows.
\end{enumerate}
\section{Related Work}
\label{sec:related}

\paragraph{Multi-agent systems with LLMs.}
A first wave of LLM multi-agent work hand-designs roles, prompts, and coordination protocols for specific applications. AutoGen~\citep{autogen}, MetaGPT~\citep{metagpt}, and CAMEL~\citep{camel} provide programming abstractions for orchestrating multiple specialised LLM agents and demonstrate strong results on software engineering, research, and embodied tasks. Single-agent prompting techniques such as chain-of-thought~\citep{cot}, tree-of-thoughts~\citep{tot}, self-refine~\citep{selfrefine}, multi-persona collaboration~\citep{multipersona}, ReAct~\citep{react}, and Voyager~\citep{voyager} are special cases of multi-agent design with one or two roles. While effective, these systems require human experts to choose the agent set and the coordination structure for each new task, limiting scalability and adaptability.

\paragraph{Automated MAS generation at the task level.}
A second wave of work automates parts of the design process. AFlow~\citep{aflow} formulates workflow optimisation as a search problem over code-represented operator graphs and uses Monte Carlo Tree Search to discover effective workflows from a fixed operator vocabulary. ADAS~\citep{adas} uses a meta-agent that iteratively programmes and archives new agent designs in code, leveraging the Turing-completeness of the search space. GPTSwarm~\citep{gptswarm} casts multi-agent construction as an optimisation over computational graphs. MaAS~\citep{maas} searches over a multi-agent supernet to identify high-performance subnetworks, and MetaAgent Search~\citep{metaagent} learns a controller that emits the best architecture per benchmark. FlowReasoner~\citep{flowreasoner} also operates at this granularity. All of these methods optimise a single workflow per task category and apply it uniformly to every query in that category. We argue (\S\ref{sec:experiments}) that this assumption breaks under heterogeneous workloads: query difficulty varies within a task and a single workflow cannot serve a mixed query stream that contains code, math, and QA.

\paragraph{Query-level adaptation.}
Two earlier works move toward query-level adaptation. FlowReasoner~\citep{flowreasoner} produces a workflow per query but draws agents from a fixed predefined set; the workflow is therefore restricted to the expressivity of those agents. Self-refine~\citep{selfrefine} and tree-of-thoughts~\citep{tot} adapt the inference-time computation per query but use a single underlying agent. OOPMAS combines these two ideas: it generates both the agent set and the workflow per query, and the agents are themselves dynamically synthesised from a meta-agent rather than chosen from a closed library.

\paragraph{Skill libraries and self-improvement.}
Voyager~\citep{voyager} accumulates a library of executable skills across game episodes and reuses them at later episodes. ADAS's archive plays a similar role at coarse granularity. Our skill library is closer in spirit to Voyager but stores structured failure/fix records rather than executable code, and is consumed by the meta-agent as in-context optimisation memory rather than as a callable library. Crucially, skill construction in OOPMAS is training-free: the skill library improves the meta-agent's prompt across rounds without any model parameter updates.

\paragraph{Benchmarks for MAS evaluation.}
Existing MAS-generation methods are typically evaluated on individual benchmarks (HumanEval~\citep{humaneval}, MBPP~\citep{mbpp}, MATH~\citep{math}, GSM8K~\citep{gsm8k}, HotpotQA~\citep{hotpotqa}, DROP~\citep{drop}). We argue this single-benchmark evaluation does not test the realistic deployment setting. The Mixed and Hard Mixed benchmarks introduced in this paper include heterogeneous query types and output formats per run, exposing failure modes of task-level methods that single-benchmark evaluation hides. The Hard Mixed benchmark additionally tests competition-level difficulty (AIME, LiveCodeBench~\citep{}, CodeContests~\citep{codecontest}, GPQA~\citep{gpqa}) that has not yet saturated, providing headroom to distinguish methods.

\section{Method}

\begin{figure}[t]
\centering
\includegraphics[width=\linewidth]{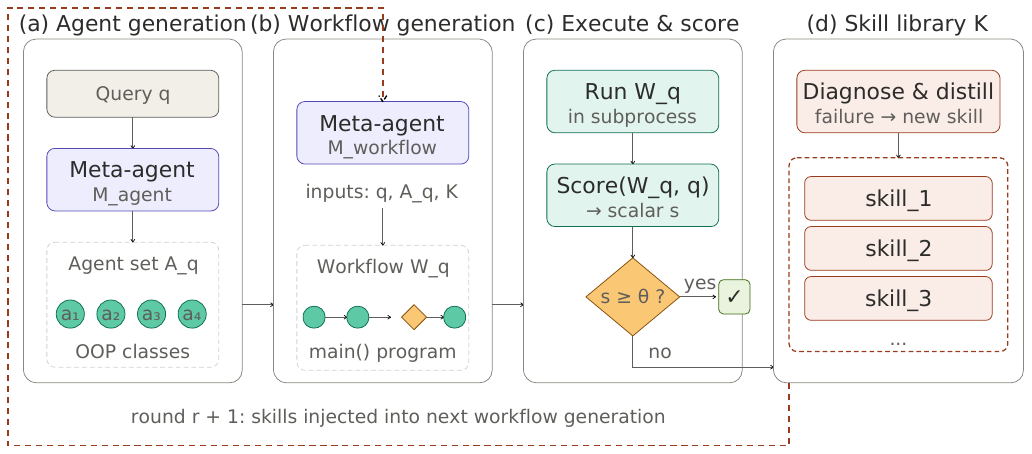}
\caption{\textbf{The OOPMAS pipeline.} \textbf{(a)}~A meta-agent emits a
tailored agent set $A_q$ for each query. \textbf{(b)}~A second meta-agent
emits an executable workflow $W_q$ over $A_q$, conditioned on the skill
library $K$. \textbf{(c)}~$W_q$ is run and scored; failures are diagnosed.
\textbf{(d)}~Each diagnosis distills a reusable skill into $K$, which is
injected into workflow generation in subsequent rounds. Training-free.}
\label{fig:oopmas-pipeline}
\end{figure}

In this section, we first formalize the query-level MAS generation problem in section~\ref{sec:prob-form}, then we describe the two core phases of OOPMAS: dynamic agent generation and workflow generation and optimization. Figure~\ref{fig:oopmas-pipeline} summarises the pipeline.
\subsection{Problem Formulation}
\label{sec:prob-form}
\paragraph{Multi-Agent System.}
We define a multi-agent system (MAS) for a query $q$ as a tuple $S_q = (\mathcal{A}_q, \mathcal{W}_q)$, where $\mathcal{A}_q = \{a_1, \ldots, a_n\}$ is a set of agents and $\mathcal{W}_q$ is a workflow that coordinates them. Each agent $a_i = (p_i, \mathcal{T}_i, m_i(h_i), s_i)$ is characterized by a role prompt $p_i$ specifying its role and expertise, a tool set $\mathcal{T}_i$ defining its action space, a backbone model $m_i$ with hyperparameter such as temperature $h_i$, and states $s_i$ that contains conversation history and intermediate results. The workflow $\mathcal{W}_q$ is an executable program over $\mathcal{A}_q$.

\paragraph{Query-Level Workflow Optimization.}
Given a query $q$ drawn from an arbitrary distribution $\mathcal{Q}$ and a task-specific evaluation function $\mathrm{Score}(\cdot)$, the MAS generation problem is to find:
\begin{equation}
\label{eq:objective}
S_q^* = \arg\max_{S_q \in \mathcal{S}} \; \mathrm{Score}(S_q, q),
\end{equation}
where $\mathcal{S} = \{(\mathcal{A}, \mathcal{W}) : \mathcal{A} \subseteq \mathcal{A}^*, \, \mathcal{W} \text{ over } \mathcal{A}\}$ is the space of all valid multi-agent systems and $\mathcal{A}^*$ is the space of all possible agent configurations. This is a per-query optimization: the system $S_q^*$ is generated fresh for each $q$, allowing both the agent composition and the workflow topology to adapt to the specific demands of the query.

\subsection{Dynamic Agent Generation}
\label{sec:agent-gen}
We represent both agents and workflows as executable Python code: agents follow an object-oriented programming (OOP) convention and the corresponding workflow is a main function that imports agent classes.

\paragraph{OOP Agent Representation.}
Each agent in OOPMAS is a self-contained Python object whose attributes encode the role prompt $p_i$, model configuration $m_i(h_i)$, and persistent state $s_i$, and whose methods expose the tool set $\mathcal{T}_i$ and predefined tasks. This design lets an agent accumulate conversation history and intermediate results in $s_i$ across multiple interactions inside a single workflow execution, going beyond a single LLM API call. We choose OOP for three reasons: it naturally encapsulates per-agent state without external bookkeeping; it lets inter-agent interactions be expressed as flexible Python object interactions (passing objects as arguments, reading exposed state, calling each other's methods) rather than being constrained to a fixed workflow schema; and OOP code is well represented in LLM pretraining corpora, making workflow code generation more reliable than custom DSLs or graph schemas.

\paragraph{Agent Pool Generation.}
Given validation queries $\{q_1, \ldots, q_N\}$ from a task distribution, the meta-agent $\mathcal{M}_{\mathrm{agent}}$ generates a shared agent pool $\bar{\mathcal{A}}$ in a single call. The pool is regenerated per task set; unlike methods that rely on a fixed operator set~\citep{aflow}, the roles and expertise in $\bar{\mathcal{A}}$ are tailored to the query distribution.

\paragraph{Per-Query Coverage Verification.}
Although $\bar{\mathcal{A}}$ targets common capabilities across queries, individual queries may need specialised expertise absent from the pool. For each $q_i$ we run a coverage check: the system inspects $\bar{\mathcal{A}}$ together with $q_i$, and if the pool lacks a required capability it generates additional specialist agents and merges them into a query-specific set $\mathcal{A}_{q_i}$. This two-tier process balances pool reuse with per-query adaptability.

\subsection{Workflow Generation and Optimisation}
\label{sec:workflow-opt}

Given the agent set $\mathcal{A}_q$, OOPMAS generates a workflow $\mathcal{W}_q$ and iteratively refines it over $R$ optimisation rounds. Rather than training the meta-agent via RL or distillation, we keep the meta-agent fixed and pursue cross-query improvement through a dynamic skill library $\mathcal{K}$ that accumulates structured experience from execution feedback. The pipeline is therefore training-free: no gradient updates, distillation, or fine-tuning are required, yet workflow generation improves across rounds.

\paragraph{Workflow generation and execution.}
In each round, $\mathcal{M}_{\mathrm{workflow}}$ receives $q$, $\mathcal{A}_q$, and $\mathcal{K}$, and generates an executable main function that imports agent classes, calls their interfaces, invokes tool functions, and composes outputs into a final answer. Because the workflow is unrestricted Python code over OOP agent objects, the meta-agent can express arbitrary coordination patterns without being limited to predefined operators. A lightweight static check rejects programmes that are syntactically invalid, lack an entry point, or never call \texttt{agent.run}; the meta-agent regenerates with the error messages, up to $K$ repair attempts. The validated workflow runs in a standalone subprocess and is scored by a task-specific function $\mathrm{Score}$.

\paragraph{Optimisation via the skill library.}
After each round we classify outcomes into \emph{workflow crash}, \emph{wrong output}, or \emph{success}. For crashes and wrong outputs, the meta-agent receives the failing workflow, the error trace, and $\mathcal{K}$, and produces a structured diagnosis (root cause, fix summary, fix reasoning) followed by a corrected workflow that is re-executed. The skill library $\mathcal{K}$ is a structured store of \texttt{FailureRecord} and \texttt{FixRecord} entries, organised into four prompt-injectable sections: round-level score trajectories, crash patterns by error type, successful root-cause/fix pairs, and failed fix attempts. The full skill schema and a worked example are given in Appendix~\ref{app:skills}. Algorithm~\ref{alg:oopmas} summarises the full pipeline; Phase~2 runs all queries in parallel and accumulates experience across rounds.

\begin{algorithm}[t]
\caption{OOPMAS: Training-Free Query-Level MAS Generation}
\label{alg:oopmas}
\begin{algorithmic}[1]
\REQUIRE Queries $\{q_1, \ldots, q_N\}$, evaluation function $\mathrm{Score}$, rounds $R$
\ENSURE Optimized workflows $\{\mathcal{W}_{q_i}^*\}_{i=1}^N$
\STATE \textbf{Phase 1: Agent Generation}
\STATE $\bar{\mathcal{A}} \leftarrow \mathcal{M}_{\mathrm{agent}}(\{q_1, \ldots, q_N\})$ \hfill $\triangleright$ Pooled generation (single LLM call)
\FOR{each query $q_i$}
    \STATE $\mathcal{A}_{q_i} \leftarrow \textsc{CoverageCheck}(\bar{\mathcal{A}}, q_i)$ \hfill $\triangleright$ Augment if needed
\ENDFOR
\STATE \textbf{Phase 2: Iterative Workflow Optimization}
\STATE Initialize skill library $\mathcal{K} \leftarrow \emptyset$
\FOR{round $r = 1$ to $R$}
    \FOR{each query $q_i$ \textbf{in parallel}}
        \STATE $\mathcal{W}_{q_i}^{1 \ldots N} \leftarrow \mathcal{M}_{\mathrm{workflow}}(q_i, \mathcal{A}_{q_i}, \mathcal{K})$ \hfill $\triangleright$ Generate workflow
        \STATE $s_i^{1 \ldots N} \leftarrow \mathrm{Score}(\mathcal{W}_{q_i}^{1 \ldots N}, q_i)$ \hfill $\triangleright$ Execute and evaluate
        \FOR{ $ w_{q_i}^{j} \in \mathcal{W}_{q_i}^{1 \ldots N} $} 
            \IF{$s_i^j \in s_i^{1 \ldots N} < \theta$}
                \STATE $\mathcal{W}_{q_i}' \leftarrow \textsc{Diagnose\&Fix}(\mathcal{W}_{q_i}, \mathrm{error}_i, \mathcal{K})$
                \STATE $s_i' \leftarrow \mathrm{Score}(\mathcal{W}_{q_i}', q_i)$
            \ENDIF
        \ENDFOR
    \ENDFOR
    \STATE $\mathcal{K} \leftarrow \mathcal{K} \cup \{\text{failures, fixes, scores from round } r\}$
\ENDFOR
\end{algorithmic}
\end{algorithm}

% ============================================================
% OOPMAS Experiment Section — NeurIPS 2025 format
% ============================================================

\section{Experiments}
\label{sec:experiments}

We evaluate OOPMAS on four experimental settings: (i)~a \emph{mixed-task} benchmark of standard tasks (\S\ref{sec:exp-mixed}); (ii)~a \emph{hard mixed-task} benchmark of competition-level tasks (\S\ref{sec:exp-mixed-hard}) to test generalisation as task difficulty increases; (iii)~a component ablation that isolates the contributions of the agent pool, skill library, and optimisation memory (\S\ref{sec:exp-ablation}); and (iv)~a model-swap study together with a cost--accuracy comparison against baselines (\S\ref{sec:exp-model-swap},~\S\ref{sec:exp-cost}). All experiments use GPT-5 unless stated otherwise.

\subsection{Experimental Setup}
\label{sec:exp-setup}

\paragraph{Mixed-Task Benchmark.}
Prior automated MAS-generation methods evaluate each task type separately: AFlow~\citep{aflow} and ADAS~\citep{adas} run a separate optimisation per benchmark for HumanEval~\citep{humaneval}, MBPP~\citep{mbpp}, MATH~\citep{math}, GSM8K~\citep{gsm8k}, HotpotQA~\citep{hotpotqa}, and DROP~\citep{drop}, producing one optimised workflow per benchmark. This task-specific generation does not reflect real-world deployment: in practice, users submit queries from a mixture of task types with unpredictable proportions, and the MAS-generation system must handle them with a single pipeline. To evaluate this setting, we construct a mixed-task benchmark of $120$~queries: $40$~code (HumanEval, MBPP), $40$~math (MATH, GSM8K), and $40$~QA (HotpotQA, DROP). All samples are drawn from the validation splits.

\paragraph{Hard Mixed-Task Benchmark.}
The standard mixed benchmark is now largely saturated by strong frontier LLMs, so accuracy alone no longer separates methods. To stress-test query-level adaptation under realistic difficulty, we additionally construct a \emph{hard mixed-task} benchmark of $180$~queries drawn from competition-level sources: $60$~math from AIME~2025/2026, $60$~code from LiveCodeBench~\citep{livecodebench} and CodeContests~\citep{codecontest}, and $60$~QA from GPQA~Diamond~\citep{gpqa}. Each domain has heterogeneous formats (integer answers for AIME, full \texttt{stdin}/\texttt{stdout} programs for CodeContests, four-way multiple choice for GPQA), forcing the MAS-generation system to produce specialised workflows per query rather than a single template.

\paragraph{Baselines.}
We compare against four automated MAS-generation methods: (1)~\textbf{AFlow}~\citep{aflow}, MCTS-based optimisation of code-represented workflows over a fixed operator vocabulary; (2)~\textbf{ADAS}~\citep{adas}, a meta-agent that iteratively programmes and archives agent designs; (3)~\textbf{MaAS}~\citep{maas}, search over a multi-agent supernet; and (4)~\textbf{MetaAgent}~\citep{metaagent}. For all baselines, the system receives the same mixed batch and must produce workflows that handle every query type, matching the setting OOPMAS operates in.

\paragraph{Evaluation Metrics.}
We use pass@1 for code (HumanEval, MBPP, LiveCodeBench, CodeContests), solve rate for math (MATH, GSM8K, AIME), and F1 for QA (HotpotQA, DROP), and accuracy for multiple-choice (GPQA). We report per-domain accuracy averaged over the sub-benchmarks within each domain, and an Overall column averaged across all queries.

\paragraph{Implementation Details.}
Unless stated otherwise, all methods use GPT-5 as both the meta-agent and the worker model. OOPMAS runs for $R=5$ optimisation rounds on the mixed benchmark and $R=3$ on the hard mixed benchmark, with concurrency $30$ and at most $7$~agents per query. We report the highest score across the two configurations of OOPMAS (with the dynamic skill library on or off); other hyperparameters are reported in Appendix~\ref{app:repro}.

%% ----------------------------------------------------------------
\subsection{Mixed-Task Benchmark Results}
\label{sec:exp-mixed}

Table~\ref{tab:mixed-results} reports results on the standard mixed-task benchmark. OOPMAS reaches $89.6\%$ overall accuracy and outperforms every baseline by a wide margin: the best baseline, ADAS, reaches $71.5\%$, an absolute gap of $18.1$ percentage points. The advantage is largest on code generation (+$26.0$ points over ADAS) and persists across math and QA. Task-level baselines exhibit characteristic failure modes on heterogeneous distributions. AFlow's MCTS optimisation discovers a single workflow specialised for one domain; that workflow scores $90.0$ on code but $0.0$ on math and QA, collapsing the overall score to $30.0$. MaAS exhibits the same pathology in reverse, achieving reasonable math accuracy but near-zero on code. ADAS and MetaAgent are more balanced because their archives include heterogeneous solutions, but they still under-perform OOPMAS by a wide margin. OOPMAS's per-query generation tailors both the agent set and the workflow topology to the specific demands of each query, so simple problems receive a single-expert call while harder ones receive multi-agent pipelines with planners, reviewers, and iterative testing loops.

\begin{table}[t]
\caption{Accuracy ($\%$) on the \textbf{Mixed} and \textbf{Hard Mixed} benchmarks. All methods use GPT-5 as the worker model and receive the same mixed batch of queries. Best results in \textbf{bold}.}
\label{tab:mixed-results}
\centering
\small
\setlength{\tabcolsep}{3.5pt}
\begin{tabular}{lcccccccc}
\toprule
& \multicolumn{4}{c}{\textbf{Mixed} }
& \multicolumn{4}{c}{\textbf{Hard Mixed}} \\
\cmidrule(lr){2-5} \cmidrule(lr){6-9}
Method & Code & Math & QA & Overall & Code & Math & QA & Overall \\
\midrule
ADAS & 50.0 & 82.5 & 82.0 & 71.5 & --- & --- & --- & 72.3 \\
MetaAgent & 50.0 & 80.0 & 80.7 & 70.2 & 64.8 & 45.0 & 58.4 & 61.8 \\
MaAS$^\dagger$ & 2.5 & 77.5 & --- & 40.0 & --- & --- & --- & --- \\
AFlow & 90.0 & 0.0 & 0.0 & 30.0 & 0.0 & 20.0 & 36.5 & 8.9 \\
\midrule
OOPMAS (ours) 
& \textbf{97.5} & \textbf{85.0} & \textbf{86.4} & \textbf{89.6}
& \textbf{74.5} & \textbf{98.3} & \textbf{98.3} & \textbf{90.8} \\
\bottomrule
\end{tabular}

\vspace{0.4em}
\end{table}

%% ----------------------------------------------------------------
\subsection{Hard Mixed-Task Benchmark}
\label{sec:exp-mixed-hard}

Hard Mixed (right half of Table~\ref{tab:mixed-results}) is substantially more challenging: it pairs competition-level math (AIME) with stdin/stdout competitive programming (LiveCodeBench, CodeContests) and graduate-level science MCQ (GPQA). Despite the increased difficulty, OOPMAS reaches $90.8\%$ overall, again leading the strongest baseline (ADAS, $72.3\%$) by $18.5$~percentage points -- almost identical to the gap on the easier mixed benchmark. The per-query agent and workflow generation thus generalises to harder problem distributions.

The baseline failure modes on Hard Mixed are informative. AFlow's single MCTS-discovered workflow assumed HumanEval-style function-completion outputs and is incompatible with stdin/stdout competitive programming, collapsing to $0\%$ on code. MetaAgent's code accuracy improves slightly ($50 \to 64.8$) but math drops sharply ($80 \to 45$) because the AIME integer-answer format is out of distribution. ADAS is the most consistent baseline ($71.5 \to 72.3$) but still trails by $18.5$ points. These patterns reinforce the central argument: a single fixed workflow cannot accommodate heterogeneous output formats and difficulty levels in realistic mixed workloads, while query-level generation handles them naturally. OOPMAS's edge widens on math and QA, where per-query workflows exploit problem structure (chain-of-specialists for AIME, multi-perspective consensus for ambiguous GPQA items); on code the lead is smaller because failure modes (off-by-one, edge cases) are orthogonal to workflow topology.

%% ----------------------------------------------------------------
\subsection{Component Ablation}
\label{sec:exp-ablation}

\begin{figure}[t]
\centering
\includegraphics[width=0.85\linewidth]{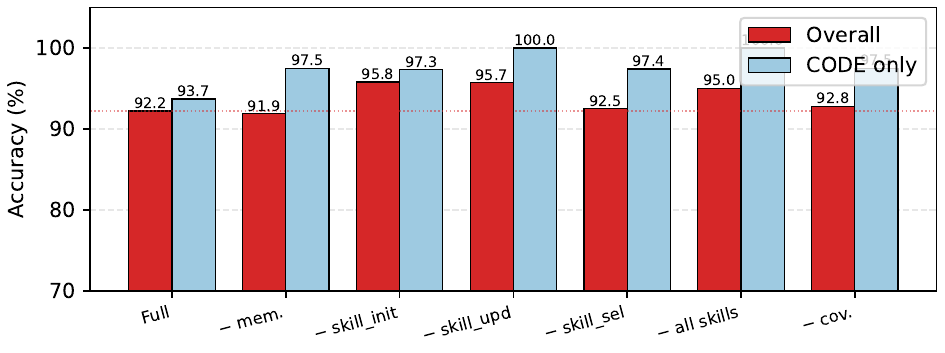}
\caption{\textbf{Component ablation on Mixed.} Removing each component individually leaves overall accuracy within $\pm 3.5$ points of Full (dotted red line). The agent pool and per-query generation are the dominant accuracy drivers; skill-library and memory components contribute marginal additional points on the saturated Mixed-validation set.}
\label{fig:ablation}
\end{figure}

We ablate the four post-pool components of OOPMAS on the Mixed validation set and report results in Figure~\ref{fig:ablation}. Each ablation removes exactly one component while keeping the others enabled. \textit{Full} corresponds to the median of four random-seed reruns of the complete system. Removing the optimisation memory~($-$mem.), skill creation~($-$skill\_init), skill updates~($-$skill\_upd), skill selection~($-$skill\_sel), all three skill components together~($-$all skills), or per-query coverage verification~($-$cov.) leaves overall accuracy within a $\pm 3.5$-point band of the Full configuration. Code accuracy is the most stable across ablations, ranging from $93.7$ to $100.0$. The key takeaway is robustness: the OOP agent pool generated for the query distribution already captures most of the headroom on Mixed, and the optimisation components matter most when the validation distribution is small or when test problems differ from validation problems. Crucially, every ablation still beats every baseline in Table~\ref{tab:mixed-results} by at least $20$ points.

%% ----------------------------------------------------------------
\subsection{Model Swap Study}
\label{sec:exp-model-swap}

\begin{figure}[t]
\centering
\includegraphics[width=0.78\linewidth]{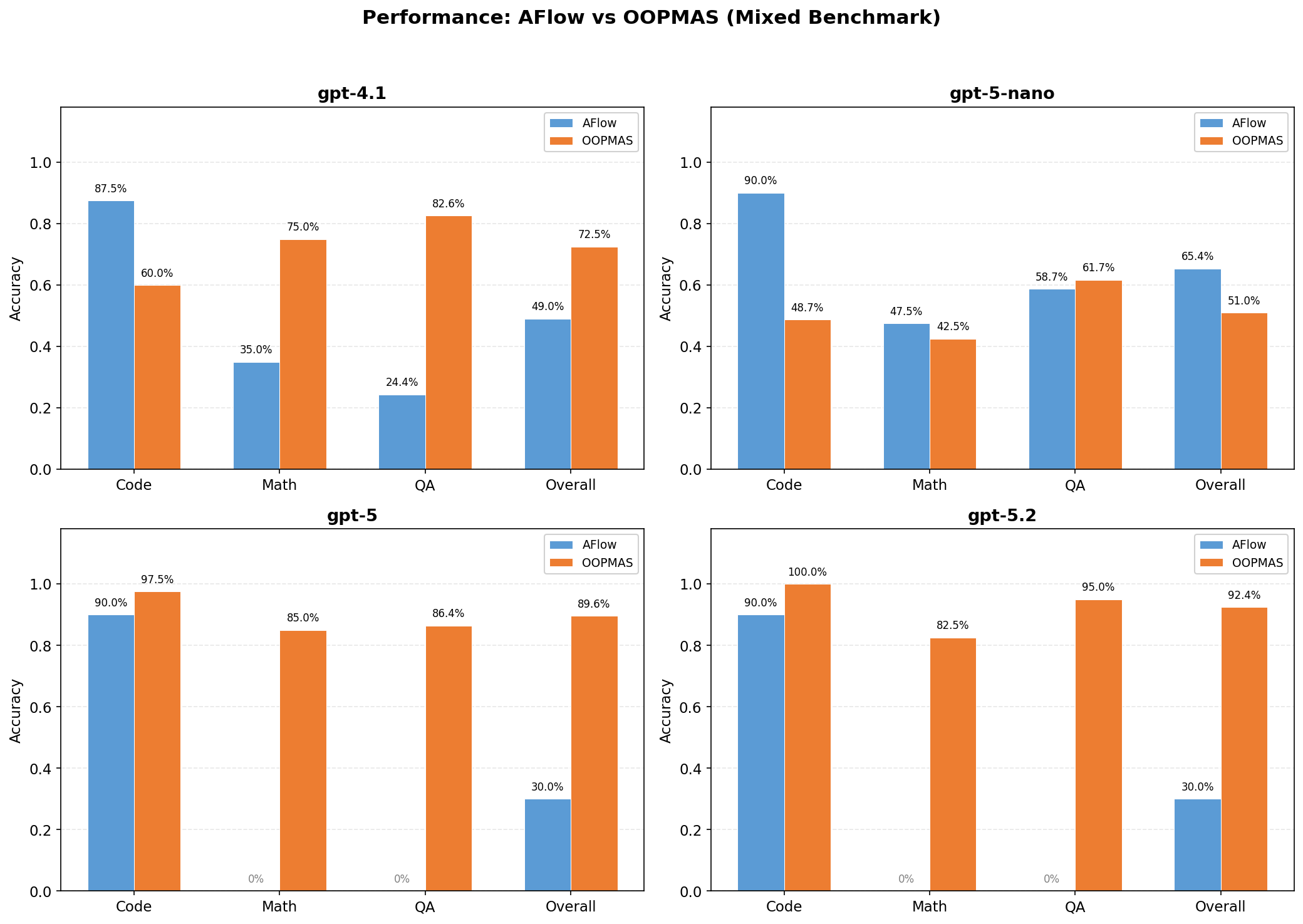}
\caption{\textbf{Model-swap study on Mixed.} OOPMAS scales consistently with the worker model's capability ($51.0 \to 92.4$ from GPT-5-nano to GPT-5.2). At weak models (GPT-5-nano), AFlow's fixed MCTS scaffold is more robust because it does not require the meta-agent to author high-quality agent code; at stronger models, dynamic generation dominates.}
\label{fig:model-swap}
\end{figure}

We re-run OOPMAS and AFlow on the Mixed benchmark with four different worker models -- GPT-4.1, GPT-5-nano, GPT-5, and GPT-5.2 -- regenerating agents and workflows from scratch each time. Figure~\ref{fig:model-swap} shows the per-domain breakdown. OOPMAS consistently outperforms AFlow on GPT-4.1, GPT-5, and GPT-5.2, and scales smoothly with model capability ($72.5\% \to 92.4\%$ from GPT-4.1 to GPT-5.2). The exception is GPT-5-nano, where AFlow's fixed MCTS workflow ($65.4\%$) outperforms OOPMAS ($51.0\%$). With a weak meta-agent, the per-query agent code is too unreliable; AFlow's pre-discovered scaffold is more robust because it does not require online code synthesis. As the worker model strengthens, dynamic generation rapidly dominates, exceeding the fixed workflow by $\approx27$ points at GPT-5.2.

%% ----------------------------------------------------------------
\subsection{Cost--Accuracy Trade-off}
\label{sec:exp-cost}

\begin{figure}[t]
\centering
\includegraphics[width=0.92\linewidth]{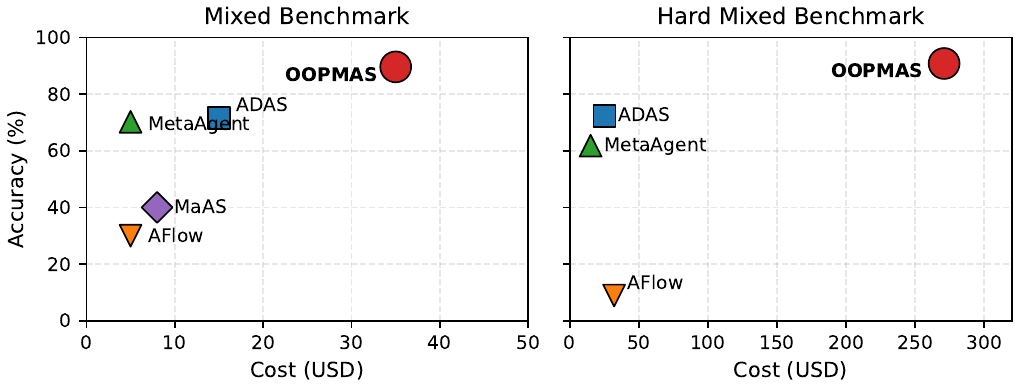}
\caption{\textbf{Accuracy versus cost} on Mixed (left) and Hard Mixed (right). OOPMAS sits in the upper-right of both panels. Costs cover the full pipeline; baseline costs are estimates derived from the configuration used in Table~\ref{tab:mixed-results}.}
\label{fig:cost}
\end{figure}

Figure~\ref{fig:cost} plots accuracy against total dollar cost. On Mixed, OOPMAS uses $\$35$ to reach $89.6\%$ vs.\ $\$15$ for ADAS at $71.5\%$ -- a $\sim\!7\times$ premium for $+18$ accuracy points. On Hard Mixed, OOPMAS at $\$271$ reaches $90.8\%$ vs.\ ADAS at $\$25$ for $72.3\%$. The cost premium buys \emph{robustness} (OOPMAS is the only method that maintains accuracy as difficulty grows) rather than marginal accuracy, and the per-query nature of the cost makes selective deployment easy: route cheap queries to a single-expert workflow and expensive queries to the full pipeline. The framework remains training-free, so deploying on a new domain only requires changing the validation queries. A per-stage cost decomposition is in Appendix~\ref{app:cost}.

\section{Conclusion}
\label{sec:conclusion}

\section{Conclusion}

We presented OOPMAS, a training-free framework for query-level multi-agent system generation. Unlike prior automated MAS design methods that optimize a single fixed workflow for an entire task or benchmark, OOPMAS treats each query as its own system-design problem. For every query, it jointly synthesises a tailored set of object-oriented agents and an executable coordination workflow, then refines the workflow over $R$ optimization rounds using execution feedback and a dynamic skill library. This design expands the search space from workflow-only optimization over a fixed agent pool to joint agent-and-workflow generation, allowing the system to adapt both the expertise available to the workflow and the structure of the workflow itself.

\paragraph{Limitation.} Several limitations remain. First, OOPMAS depends on a strong meta-agent. When the backbone is weak, the system may fail to synthesize reliable agent classes or executable workflows, and fixed scaffolds such as AFlow can be more robust. Second, the current skill library is run-local: it improves workflow generation within an optimization run but does not yet persist across deployments, domains, or future query streams. Third, per-query generation increases inference cost substantially compared with fixed-workflow baselines, especially on hard problems that require multiple repair attempts. Finally, while Mixed and Hard Mixed cover diverse domains and formats, they remain benchmark-based evaluations; broader deployment settings may introduce tool-use, long-context, safety, privacy, and distribution-shift challenges not fully captured here.

\clearpage
\bibliographystyle{plainnat}
\bibliography{Qi_MAS}
%%%%%%%%%%%%%%%%%%%%%%%%%%%%%%%%%%%%%%%%%%%%%%%%%%%%%%%%%%%%
\clearpage
\appendix

\section*{Technical Appendices }
% =====================================================================
% Appendix
% =====================================================================
\section{Reproducibility Details}
\label{app:repro}

\paragraph{Hardware and runtime.}
Experiments were run on a single workstation with $48$~CPU threads; all LLM calls are offloaded to the OpenAI API. A typical Mixed run completes in $30$--$60$ minutes; a typical Hard Mixed run takes $1.5$--$3$ hours. Wall-clock is dominated by API latency rather than local compute.

\paragraph{Evaluation specifics.}
For HumanEval and MBPP we treat a query as solved iff every assertion in the dataset's test list passes, evaluated in an isolated subprocess. For MATH and GSM8K we use the standard \texttt{boxed} answer extraction. For HotpotQA and DROP we use F1 on the predicted answer. For AIME we extract the final integer in the range $[0, 999]$ and compare to ground truth. For LiveCodeBench and CodeContests we run the predicted programme on every public and private test case and require all to pass. For GPQA we extract the final letter $\in \{A, B, C, D\}$.

% ---------------------------------------------------------------------
\section{Per-Stage Cost Breakdown}
\label{app:cost}

\begin{figure}[h]
\centering
\includegraphics[width=0.85\linewidth]{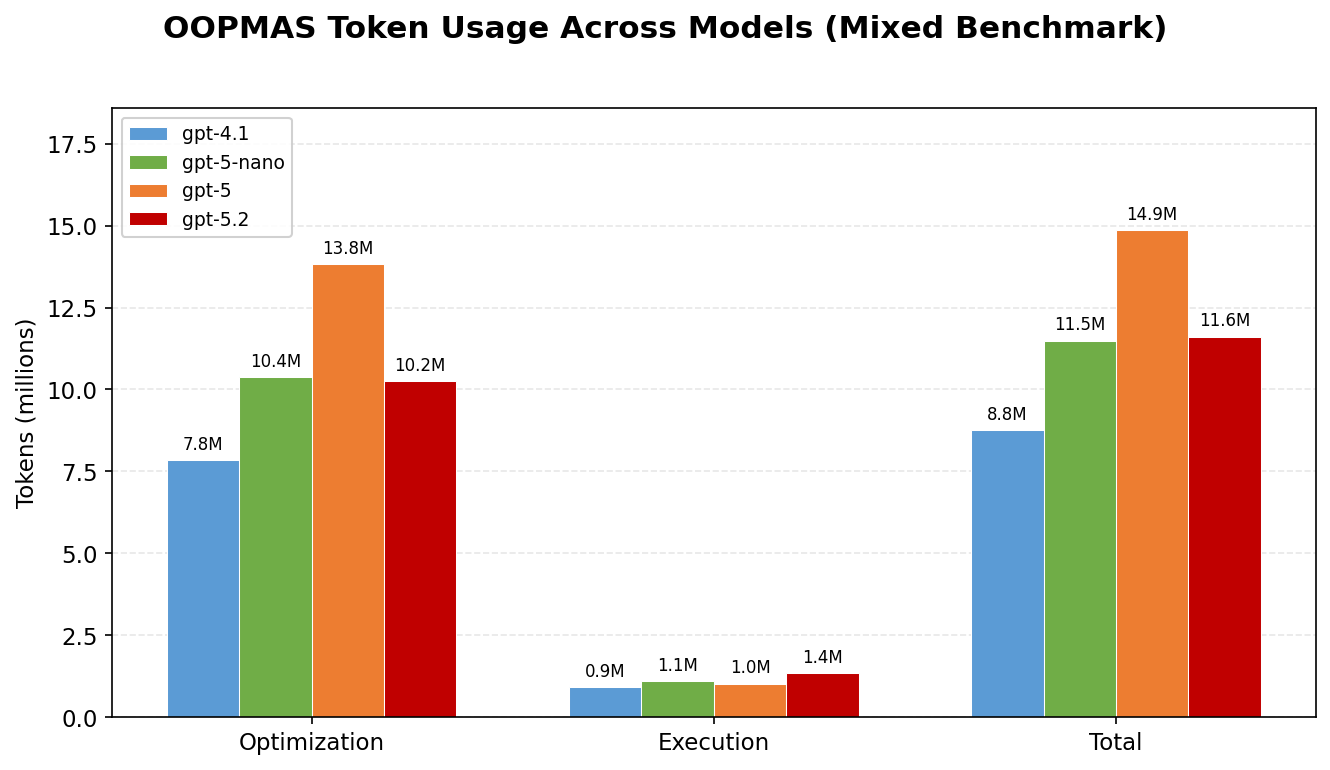}
\caption{Token usage by OOPMAS on the Mixed benchmark, decomposed into Optimisation (workflow generation, skill update, repair) and Execution (per-agent inference). Total tokens scale roughly linearly with model size up to GPT-5; GPT-5.2 uses fewer optimisation tokens than GPT-5 because its workflows succeed earlier and require less repair.}
\label{fig:tokens}
\end{figure}

Figure~\ref{fig:tokens} reports the total token usage for OOPMAS across four backbone models. The Optimisation phase dominates (workflow generation, repair, skill creation), while Execution (per-agent calls inside the workflow) is small in comparison because most queries terminate within a single agent call. A GPT-5 run on Mixed costs approximately \$$35$ at current public API pricing, or \$$0.30$ per query. On Hard Mixed, the per-query cost rises to \$$1.50$ because problems are longer and require more repair attempts. Within a single Mixed run, the per-stage decomposition is approximately: agent generation $0.3\%$, skill creation/update $1.0\%$, skill selection $4.5\%$, workflow generation $58\%$, workflow execution $36\%$ (numbers from a representative GPT-5 run; full breakdowns are produced by the codebase under each run's \texttt{cost\_summary.json}).

% ---------------------------------------------------------------------
\section{Skill Library Examples}
\label{app:skills}

The skill library is a structured store of \texttt{FailureRecord} and \texttt{FixRecord} entries serialised between rounds. A single skill manifest entry has the schema:

\begin{verbatim}
{
  "skill_id": "code_generate_test_refine",
  "name": "Code: Generate-Test-Refine Loop",
  "applicable_domains": ["code"],
  "when_to_use": "Code-generation problems with verifiable test cases.",
  "strategy": "1. Pass the full task to a code-writer agent.
               2. Run the function's tests via execute_code.
               3. On failure, pass error+code to a code-refiner agent.
               4. Print the final code between markers.",
  "detection_keywords": ["writer.run", "execute_code", "CODE_START"]
}
\end{verbatim}

After each optimisation round, the meta-agent reviews up to $30$ failure traces and $10$ successful fixes from the round and produces an updated manifest. We constrain each skill to a single domain to prevent the workflow generator from importing cross-domain agents into a single workflow (e.g., importing a math agent into a code workflow), which is a major source of contamination in early prototypes.

\end{document}